\pdfoutput=1

\documentclass[11pt]{article}

\usepackage[]{acl}

\usepackage{times}
\usepackage{latexsym}

\usepackage[T1]{fontenc}
\usepackage{amsmath}
\usepackage{listings}
\usepackage[utf8]{inputenc}

\usepackage{microtype}

\usepackage{times}
\usepackage{amssymb}
\usepackage{latexsym}
\usepackage{booktabs}
\usepackage{xspace}
\usepackage{xcolor}
\usepackage{todonotes}
\usepackage{graphicx}
\usepackage{todonotes}
\usepackage{xspace}
\usepackage{hyperref}
\usepackage{multirow}
\usepackage{rotating}
\usepackage[super]{nth}

\definecolor{royalpurple}{RGB}{136,18,255}
\definecolor{royalblue}{RGB}{0,102,204}

\newcommand{\paradocs}{\textsc{ParaDocs}\xspace}
\newcommand{\contexts}{\textsc{Docs}\xspace}
\newcommand{\sentences}{\textsc{Sents}\xspace}
\newcommand{\loose}{\textsc{Loose\textsubscript{75\%}}\xspace}
\newcommand{\medium}{\textsc{Medium\textsubscript{50\%}}\xspace}
\newcommand{\strict}{\textsc{Strict\textsubscript{25\%}}\xspace}

\title{Recovering document annotations for sentence-level bitext}

\author{Rachel Wicks$^{1, 2}$, Matt Post$^{1-3}$, Philipp Koehn$^{1,2}$\\
  ${}^1$Human Language Technology Center of Excellence, Johns Hopkins University\\
  ${}^2$Center of Language and Speech Processing, Johns Hopkins University \\
  ${}^3$Microsoft \\
  \texttt{rewicks@jhu.edu, mattpost@microsoft.com, phi@jhu.edu}}

\begin{document}
\maketitle
\begin{abstract}

Data availability limits the scope of any given task.
In machine translation, historical models were incapable of handling longer contexts, so the lack of document-level datasets was less noticeable.
Now, despite the emergence of long-sequence methods, we remain within a sentence-level paradigm and without data to adequately approach context-aware machine translation.
Most large-scale datasets have been processed through a pipeline that discards document-level metadata.
In this work, we reconstruct document-level information for three (ParaCrawl, News Commentary, and Europarl) large datasets in German, French, Spanish, Italian, Polish, and Portuguese (paired with English).
We then introduce a document-level filtering technique as an alternative to traditional bitext filtering.
We present this filtering with analysis to show that this method prefers context-consistent translations rather than those that may have been sentence-level machine translated.
Last we train models on these longer contexts and demonstrate improvement in document-level translation without degradation of sentence-level translation.
We release our dataset, \paradocs, and resulting models as a resource to the community.\footnote{\url{https://huggingface.co/datasets/jhu-clsp/paradocs}}

\end{abstract}

\section{Introduction}
\label{sec:introduction}

Since the early days of statistical methods, machine translation has been centered within a sentence-level paradigm.
N-gram based approaches, which typically obey sentence-boundaries, were the predominant machine translation method and did not effectively use the wealth of information contained across contexts \cite{marino-etal-2006-n}.
Later, newer neural techniques, such as Transformers \cite{DBLP:journals/corr/VaswaniSPUJGKP17}, became popular and have been shown to be effective at handling longer sequences \cite{DBLP:journals/corr/abs-2004-05150,sun-etal-2022-rethinking,post2023escaping}.
During this time, researchers have periodically considered context-aware machine translation in training methodologies \cite{voita-etal-2018-context}, evaluation sets and metrics \cite{jiang-etal-2022-blonde,vernikos-etal-2022-embarrassingly,muller-etal-2018-large}, and its unique ability to address discourse phenomena which are otherwise impossible to correctly translate without context \cite{voita-etal-2019-good,bawden-etal-2018-evaluating}.
To present, little work has been done to address the most obvious hurdle: a lack of document-level training data.

\begin{figure*}[t]
    \centering
    \includegraphics[width=\textwidth]{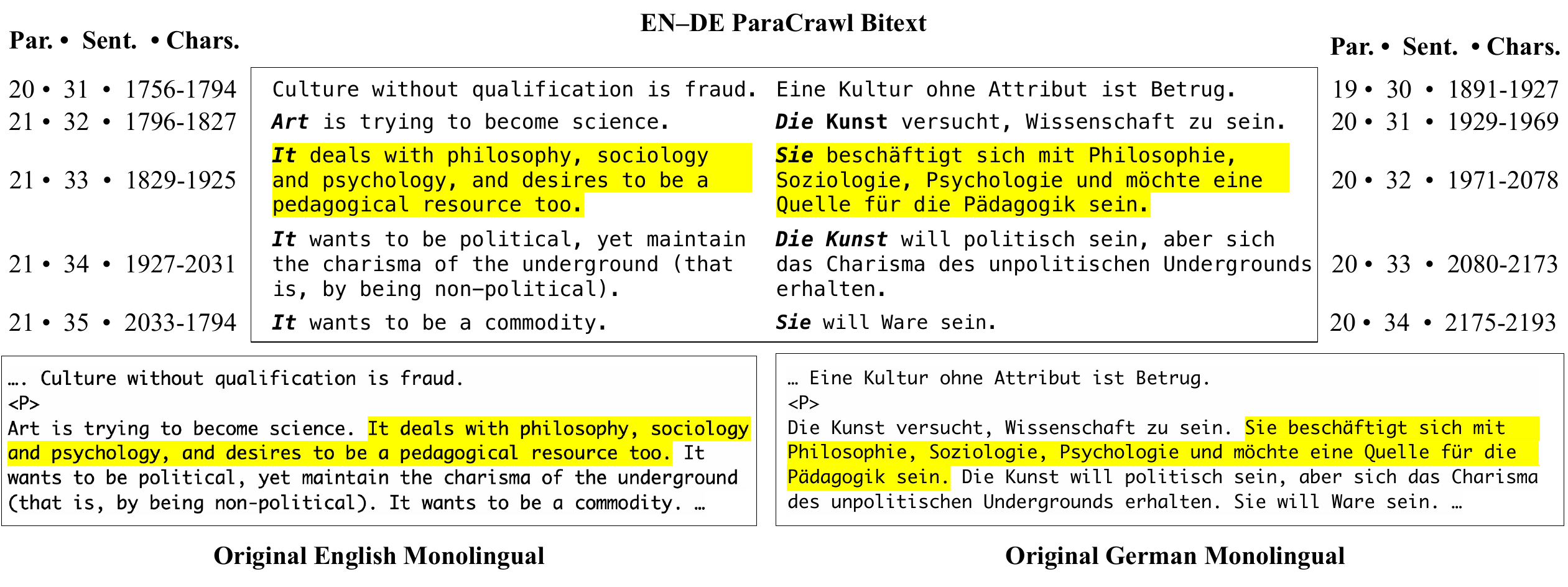}
    \caption{An example from ParaCrawl. The existing bitext has no contextual information. A model is trained to produce ``Sie'' (a feminine pronoun) from ``It'' without appropriate context. We restore this information by finding text in the corresponding monolingual dumps, and add document, paragraph, sentence, and character offset metadata.}
    \label{fig:extraction}
    \vspace{-2mm}
\end{figure*}

Many cornerstone datasets were created by finding known sources of professionally translated documents \cite{koehn-2005-europarl, kocmi-etal-2023-findings, banon-etal-2020-paracrawl}.
These parallel documents were then sentence-segmented and aligned with a document-based alignment technique.
When more data was demanded, these datasets proved insufficient, so data curators moved towards \textit{global mining}---treating web crawls as a bag of sentences and searching for the most similar sentence in the target language \cite{schwenk-etal-2021-ccmatrix,schwenk-etal-2021-wikimatrix, el-kishky-etal-2020-ccaligned}.
This removes document order and makes document-reconstruction impossible.

Some datasets, such as ParaCrawl, exist as a sentence-level resource but have been constructed in a way more amenable to document-reconstruction.
This work confronts this resource gap by providing document-level annotations for News Commentary, Europarl, and the unfiltered ParaCrawl \cite{banon-etal-2020-paracrawl} data.
Our contributions are:
\begin{enumerate}
    \setlength\itemsep{0.2em}
    \item the reconstruction of documents from three large, popular datasets, illustrated in Figure~\ref{fig:extraction};
    \item the implementation of a document-level filtering technique as an alternative to traditional bitext filtering which destroys document-level metadata;
    \item analysis that shows this filtering prioritizes context-consistent translations;
    \item results showing that models trained on this data are better at translating document-level phenomena without degrading sentence-level performance;
    \item the public release of this data, \paradocs, and these models.
\end{enumerate}

\section{Related Works}

Three areas of research are relevant to this work: sentence-level bitext mining, document-level bitext mining, and context-aware machine translation.

\paragraph{Sentence-level Mining} is the current default for most of the largest parallel datasets.
In the most recent WMT translation task \cite{kocmi-etal-2023-findings}, parallel training data (i.e., ParaCrawl or WikiMatrix \cite{banon-etal-2020-paracrawl, schwenk-etal-2021-wikimatrix}) tends to be sentence-level. 
Similarly, of the top ten corpora on OPUS \cite{4992de1b5fb34f3e9691772606b36edf, schwenk-etal-2021-ccmatrix, elkishky_ccaligned_2020, el2021xlent}, which make up over 93\% of their entire collection, eight\footnote{NLLB, CCMatrix, MultiCCAligned, ParaCrawl, XLEnt, MultiParaCrawl, LinguaTools-WikiTitles, CCAligned} are sentence-level \cite{schwenk-etal-2021-ccmatrix, 10.5555/3546258.3546365} and comprises over 72\% of the data,  with the exceptions being OpenSubtitles \cite{lison-tiedemann-2016-opensubtitles2016} and DGT\footnote{\url{https://ec.europa.eu/jrc/en/language-technologies/dgt-translation-memory}} which have rather specific domains.\footnote{These statistics current as of February 5, 2024, according to their website: \url{https://opus.nlpl.eu/}}
Many of these datasets are constructed via the global-mining technique, which indexes all sentences using some semantic-based hashing using LASER \cite{schwenk-douze-2017-learning} and aligns based on similarity matches irrespective of document-boundaries.

\paragraph{Document-level Mining} is representative of early datasets where document-alignment was already assumed and sentence-alignment could be done via simple features such as sentence lengths \cite{gale-church-1993-program}.
This technique was used to create original datasets such as Europarl \cite{koehn-2005-europarl}.
Other techniques are constrained by document order when aligning sentences \cite{varga-et-al-2005, sennrich-volk-2011-iterative, thompson-koehn-2019-vecalign}.
These have been used to create more recent corpora such as News Commentary \cite{kocmi-etal-2023-findings}.
ParaCrawl \cite{banon-etal-2020-paracrawl} initially aligns documents; however, the preprocessing before the final release destroys any of this preserved document-structure.
CCAligned \cite{elkishky_ccaligned_2020} has a middle-ground methodology which first aligns documents determined by both a combination of URL pairs and document similarity, followed by a LASER-based alignment.
CCAligned releases data in rough document-order; however there are no annotations for document boundaries or other context labeling.

Recent work has also explored using ParaCrawl as a source for document-level data due to the original alignment being constrained by documents \cite{al-ghussin-etal-2023-exploring}.
To extract documents, they extract paragraphs by using a subset of the data for which WMT released aligned document-level data in German--English as a document-alignment benchmark task.
This work diverges from this idea in considering the full ParaCrawl release which is a significantly larger portion of data and also expands the language set.

\paragraph{Context-Aware Machine Translation} concerns itself with any form of integrating additional context to translate.
There are a handful of well-studied discourse phenomena, such as gendered-pronoun translation, which are impossible to translate without incorporating this information \cite{wicks-post-2023-identifying, lopes-etal-2020-document, voita-etal-2019-good, muller-etal-2018-large, bawden-etal-2018-evaluating}.
Further, Transformers are entirely capable of incorporating longer contexts as input \cite{sun-etal-2022-rethinking}.
Thus, the field of context-aware translation has either involved modifying the input framework to better incorporate the signal of context-features \cite{lopes-etal-2020-document, tan-etal-2019-hierarchical, miculicich-etal-2018-document}, or studying alternative uses of training data \cite{post2023escaping,yu-etal-2020-better}.

The recent introduction of Large Language Models (LLMs) has also had a swath of studies that show these LLMs are better-situated for document-level machine translation as they naturally have significantly large context-windows \cite{wang-etal-2023-document-level, petrick-etal-2023-document, robinson-etal-2023-chatgpt, kocmi-etal-2023-findings}.
LLMs provide a desirable path towards document-level translation; however with their novelty and training pipeline opacity, it is still unknown what types of data elicits this performance in order to replicate on different sets of languages or smaller scales.

\section{Document Annotation and Reconstruction}

\begin{table*}[t]
    \centering
\begin{tabular}{l rr rrrr r}
\toprule

& 
& \textsc{Raw}

& \contexts
& \loose
& \medium
& \strict

& \sentences \\
\midrule

\multirow{2}{*}{DE} 
& \small{\# segs.}
& 6.16B %
& 161M %
& 127M %
& 87.9M %
& 41.9M %
& 257M %
\\
& \small{\# docs.}
& -
& 45.4M %
& 34.1M %
& 22.7M %
& 11.4M %
&  - %
\\
\midrule

\multirow{2}{*}{FR} 
& \small{\# segs.}
& 3.51B %
& 115M %
& 90.5M %
& 62.4M %
& 39.9M %
& 231M %

\\
& \small{\# docs.}
& - %
& 32.3M %
& 24.2M %
& 16.1M %
& 8.07M %
& -

\\
\midrule

\multirow{2}{*}{ES} 
& \small{\# segs.}
& 5.69B %
& 137M %
& 110M %
& 76.7M %
& 36.9M %
& 189M %

\\
& \small{\# docs.}
& - 
& 37.9M
& 28.5M
& 19.0M
& 9.5M
& -

\\
\midrule

\multirow{2}{*}{IT} 
& \small{\# segs.}
& 2.88B %
& 40.4M %
& 31.8M %
& 22.0M %
& 10.M %
& 112M %

\\
& \small{\# docs.}
& - 
& 12.8M %
& 9.58M %
& 6.38M %
& 3.19M %
& - 

\\

\midrule

\multirow{2}{*}{PL}
& \small{\# segs.}
& 1.09B %
& 23.2M %
& 17.9M %
& 12.2M %
& 5.85M %
& 34.2M %

\\
& \small{\# docs.}
& - 
& 7.55M %
& 5.66M %
& 3.77M %
& 1.89M %
& -
\\

\midrule

\multirow{2}{*}{PT}
& \small{\# segs.}
& 2.49B %
& 52.1M %
& 41.1M %
& 28.6M %
& 13.8M %
& 91.1M %

\\
& \small{\# docs.}
& - 
& 15.0M %
& 11.2M  %
& 7.5M %
& 3.7M %
& -

\\
\bottomrule
\end{tabular}
    \caption{Data sizes. \textsc{Raw} is the portion of data we attempted to align to an original monolingual document. \contexts is the portion of data that meets a minimal document criteria (at least two consecutive segments; > 0.5 langid; < 100 duplications). \loose, \medium, and \strict is the top 75\%, 50\%, and 25\% (respectively) of documents scored using SLIDE-based filtering. \# segs is the number of segments (roughly sentences) while \# docs. is the number of distinct \textit{sub-documents} created as described in Section~\ref{sec:context-extraction}.}
    \label{tab:data-stats}
\end{table*}

Released bitext files rarely have document annotations necessary to create contextual inputs for context-aware training.
In the case of Europarl, News Commentary, and ParaCrawl, document ids are released which point to the original monolingual files.
We use these monolingual files to reconstruct the original parallel document structure.

We construct annotations for each original aligned segment in the bitext.
Each segment has an individual source and target segment which receive separate annotations dependent on their original document.
Following the reconstruction process described in Section~\ref{sec:reconstruction}, we define a pipeline (Section~\ref{sec:context-extraction}) to extract contexts from this data.
Finally, we describe a novel document filtering method (Section~\ref{sec:document-filtering}) that can be used to filter out low-quality \textit{documents} instead of only low quality \textit{sentences}.
With each progression of these steps, one can whittle the dataset smaller while improving the relative quality of documents.

All filtered splits, and the original unfiltered data, are publicly released to facilitate further research into context-aware machine translation, and document-based data selection and filtering methods. 
The sizes of each dataset with the increased filtering is displayed in Table~\ref{tab:data-stats}.
We explore reconstruction as an alternative to re-alignment of the original documents since it is less computationally intensive and re-uses existing annotations from published datasets.
Future work can circumvent the need for reconstruction by preserving document metadata while data mining. 

\subsection{Reconstruction}
\label{sec:reconstruction}

If the end goal is to extract ``context'', it is only necessary to determine whether two source-target pairs are consecutive in the respective source and target documents.
Given an ordered list of segments\footnote{We use ``segment'' to refer to a given source or target which roughly equates to one sentence but may vary.} from the unfiltered bitext, and the original web-crawled document, we can align each segment to the original document via exact string matching.
Illustrated in Figure~\ref{fig:extraction}, this determines the index spans for each segment.
These indices identify where precisely the segment occurs in the original monolingual document.
This is relevant to identify the correct context as some sentences may have been left unaligned during the sentence-alignment process.
We annotate the spans for all source and target segments in our data.
For each segment in the \textsc{Raw} bitext, we list (with respect to the monolingual document):
\begin{itemize}
     \setlength\itemsep{0.2em}
    \item the paragraph index;
    \item the sentence index, determined by applying a Moses Sentence Splitter \cite{koehn-etal-2007-moses}; %
    \item the starting character index, after normalizing whitespace; %
    \item the ending character index, after normalizing whitespace; %
    \item the probability of the language id according to NLLB's fasttext LID model;
    \item the duplication count: the number of times this segment was repeated in ParaCrawl.\footnote{Only ParaCrawl has significant boilerplate text so we leave News Commentary and Europarl unannotated.}
\end{itemize}
Consecutive sentences are defined such that the starting character index is two more than the ending index of the previous segment.
For instance, in Figure~\ref{fig:extraction}, ``Art is trying to become a science.'' has a period that ends at index $1827$ and ``It'' of the following sentence starts at $1829$.
This indicates there is a single whitespace token separating these two segments at index $1828$ and they were originally consecutive; thus, they belong to the same context.

In Table~\ref{tab:data-stats}, we report the size of the data we reconstruct by \textsc{Raw}. 
This data precludes approximately one-third of the ParaCrawl data for which the monolingual source is CommonCrawl. 
The CommonCrawl monolingual data is only accessible by querying servers which make recovering metadata annotations computationally difficult with respect to network issues and latency. 
We leave this section of ParaCrawl unannotated.

\subsection{Context Extraction}
\label{sec:context-extraction}

The benefit of sentence-level bitext creation is that high-quality sentence-pairs are not discarded simply because the remaining document is poorly aligned.
This allows for much higher rates of alignment, and a larger overall dataset.
Unfiltered ParaCrawl documents (defined by a URL pair) are often poorly aligned, or have extraneous portions of documents in one or both languages.
We still want to keep a \textit{paragraph} of context even if the boilerplate text is of low quality.
To achieve this, we define \textit{document-breaking criteria}.
As we iterate over the annotated data, we accumulate context within document boundaries.
When a given segment fails to meet some criteria, we break the preceding and proceeding contexts into \textit{sub-documents}.
Consider an article which may have been loosely translated.
Some paragraphs may be literally translated while others may have been paraphrased or given additional context to make it more understandable to the audience.
In a given paragraph, when one segment is \textit{unaligned}, we break the paragraph into two \textit{sub-documents}: those preceding the unaligned segment and those proceeding it
These sub-documents can be still be linked by their parent document, but can be treated as independent for the sake of context extraction.
In our experiments, we break on three conditions:
\begin{enumerate}
    \item a segment is unaligned (minimum sub-document length is two)
    \item the language id probability is less than 0.5 (as predicted by NLLB's fasttext)
    \item the duplication count is more than 100---this divides documents on boilerplate texts which have high frequency and do not contribute much meaningful contextual information. 
\end{enumerate}
The resulting dataset is titled \contexts and statistics can be seen in Table~\ref{tab:data-stats}.

\subsection{Document Filtering}
\label{sec:document-filtering}

A key issue of web-crawled bitext is that much of it was machine translated \cite{thompson2024shocking}.
Sentence pairs that are \textit{well} translated will still circumvent sentence-level filtering methods such as LASER.
When these pairs sneak through, any resulting model will be nothing more than a mixture of distilled MT models from throughout the history of the internet.
As machine translated sentences on the internet are likely to have been translated at the sentence-level, this makes any resulting translation particularly prone to making errors when when translating context-phenomena \cite{muller-etal-2018-large, bawden-etal-2018-evaluating}.
Thus, it is necessary to filter out these errors before training.

\citet{peter-etal-2023-theres} showed that quality estimators (QEs), which notably do not require a reference, are capable of distinguishing fine-grained quality differences necessary for filtering.
Additionally, SLIDE \cite{raunak2023slide} showed that these same quality estimators can discriminate between context-consistent and context-inconsistent translations---as one might see when translating each sentence individually.
We combine these two ideas to propose a document-filtering methodology.

SLIDE works by creating a series of context-chunks by sliding a window across the document.
Each window is evaluated by a QE system and the scores are averaged in order to create a document-level score.
\citet{raunak2023slide} evaluates a combination of window and stride sizes.
They find some effectiveness starting at a minimum window size of three.
We use a window of three and a stride of one for our scoring.
We also chose to use the CometKiwi QE model \cite{rei-etal-2022-cometkiwi} as it is consistently a high performing model in these works.

The initial dataset is large, and much of it is low quality.
We rank documents with this scoring technique, and experiment with three different filtering cutoffs, top 75\%, 50\%, and 25\%, scored at the \textit{sub-document} level (as described in Section~\ref{sec:context-extraction}).
They are described as \loose, \medium, and \strict, respectively, in Table~\ref{tab:data-stats}.

\section{Source Data}
\label{sec:source-data}

To produce our dataset, we select source data from three large, publicly-available datasets that initially used document-level alignment.

\paragraph{ParaCrawl} \cite{banon-etal-2020-paracrawl} prioritizes a high-recall alignment so the original unfiltered data is orders of magnitude larger than the official release, but has an inferior quality.
For instance, ParaCrawl v9 \texttt{en-de} is approximately $278$M lines of cleaned sentences in arbitrary order, but an additional ``\texttt{RAW}'' file is available that contains nearly $10$B of uncleaned sequential sentence alignments.\footnote{Most of these segments are justifiably discarded during preprocessing due to low quality which is why the v9 release is substantially smaller.}
The official release has undergone the aforementioned sentence-level filtering (deduplication, similarity score filtering, etc) that removes the original context.
Fortunately, ParaCrawl releases an original unfiltered (\texttt{RAW}) data version.
This equates to each source–target pair, a document id, and a pointer to the original monolingual source.\footnote{The monolingual sources for two-thirds of ParaCrawl was released at \url{https://paracrawl.eu/moredata}.}

\paragraph{News Commentary} \cite{kocmi-etal-2023-findings} is a smaller, albeit cleaner newswire 
dataset released annually for WMT training data.\footnote{With special thanks to Barry Haddow who was kind enough to deliver us copies of the intermediate processing steps so we could recreate these annotations}
The released versions \textit{do} maintain document order, but documents are not labeled. %

\paragraph{Europarl} \cite{koehn-2005-europarl} was produced from the \textit{Proceedings of the European Parliament} which is obligatorily translated into a handful of European languages.
Europarl is typically n-way parallel which makes it ideal for machine translation despite the specific domain.
The most recent version (v10) releases document ids; however, it is not available in all languages. 
We produce the alignment ourselves with the accompanying tools.

\section{Experimental Design}
\label{sec:experimental-design}

In order to show the viability of constructing a dataset in such a fashion, we need to show two things: (1) a contextual model is at least as good at sentence-level translation quality as a sentence-level model 
and (2) a contextual model outperforms sentence-level models when considering context-based phenomena.
\begin{table*}[ht]
        \small
        \centering
\setlength{\tabcolsep}{6pt}
    \begin{tabular}{ll rrrrr rrrr rrrr}
    \toprule

    &   
    & \multicolumn{1}{c}{\textsc{Sent.}}
    & \multicolumn{3}{c}{\contexts}
    & \multicolumn{3}{c}{\loose}
    & \multicolumn{3}{c}{\medium}
    & \multicolumn{3}{c}{\strict} \\

    \cmidrule(lr){3-3} 
    \cmidrule(lr){4-6} \cmidrule(lr){7-9} \cmidrule(lr){10-12} \cmidrule(lr){13-15} 

    \multicolumn{2}{c}{Training Type:}
    & \multicolumn{1}{c}{snt.} 
    & \multicolumn{1}{c}{snt.}  & \multicolumn{2}{c}{context} 
    & \multicolumn{1}{c}{snt.}  & \multicolumn{2}{c}{context}
    & \multicolumn{1}{c}{snt.}  & \multicolumn{2}{c}{context}
    & \multicolumn{1}{c}{snt.}  & \multicolumn{2}{c}{context} \\
    \cmidrule(lr){3-3}
    \cmidrule(lr){4-4}
    \cmidrule(lr){5-6}
    \cmidrule(lr){7-7}
    \cmidrule(lr){8-9}
    \cmidrule(lr){10-10}
    \cmidrule(lr){11-12}
    \cmidrule(lr){13-13}
    \cmidrule(lr){14-15}

    \multicolumn{2}{c}{Inference Type:}
    & \multicolumn{1}{c}{snt.} 
    & \multicolumn{1}{c}{snt.} & \multicolumn{1}{c}{snt.}  & \multicolumn{1}{c}{ctx.} 
    & \multicolumn{1}{c}{snt.} & \multicolumn{1}{c}{snt.}  & \multicolumn{1}{c}{ctx.}
    & \multicolumn{1}{c}{snt.} & \multicolumn{1}{c}{snt.}  & \multicolumn{1}{c}{ctx.}
    & \multicolumn{1}{c}{snt.} & \multicolumn{1}{c}{snt.}  & \multicolumn{1}{c}{ctx.} \\

    \midrule
    \multirow{2}{*}{EN-DE} & \textsc{W23}
    & 39.8
    & 40.2
    & 39.1
    & 40.5
    & 40.6
    & 40.6
    & 42.1
    & 40.6
    & 40.6
    & \textbf{42.3}
    & 40.9
    & 40.0
    & 41.9
    \\
    & \textsc{Flo.}
    & 39.7
    & 39.8
    & 38.8
    & 40.5
    & 40.6
    & 40.3
    & \textbf{41.1}
    & 40.2
    & 40.2
    & 41.0
    & 40.7
    & 40.5
    & 40.7

    \\
    \midrule

    \multirow{2}{*}{EN-FR} & \textsc{W15}

    & 41.5
    & 41.6
    & 41.8
    & 42.3
    & 41.9
    & 41.8
    & 42.5
    & 41.9
    & 41.9
    & \textbf{42.9}
    & 41.9
    & 41.5
    & 42.1
    
    \\
    & \textsc{Flo.}
    
    & 51.9
    & 51.6
    & 51.5
    & 52.5
    & 51.7
    & 51.3
    & 52.1
    & 52.2
    & 52.4
    & \textbf{52.6}
    & 52.1
    & 51.0
    & 52.0
    
    \\
    \midrule

    \multirow{2}{*}{EN-ES} & \textsc{W13}
    & 36.0
    & 36.3
    & 36.2
    & 36.2
    & 36.1
    & 36.1
    & 36.1
    & \textbf{36.6}
    & 36.2
    & 36.4
    & 36.3
    & 36.2
    & 36.2

    \\
    & \textsc{Flo.}
    & 27.5
    & 27.8
    & 27.9
    & 28.1
    & 27.9
    & 28.0
    & \textbf{28.4}
    & 28.0
    & 27.8
    & 28.0
    & 28.0
    & 28.2
    & 28.3
    \\
    \midrule

    \multirow{2}{*}{EN-IT} & \textsc{W09}
    & 33.2
    & 33.3
    & 33.3
    & 33.2
    & 33.6
    & 33.4
    & 33.6
    & 33.7
    & 33.4
    & \textbf{33.8}
    & 33.2
    & 32.8
    & 33.3

    \\
    & \textsc{Flo.}
    & 29.8
    & 29.1
    & 29.2
    & 29.1
    & 29.8
    & 29.8
    & 29.8
    & \textbf{30.3}
    & 29.7
    & 29.8
    & 30.0
    & 29.9
    & 29.7
    \\
    \midrule

    \multirow{2}{*}{EN-PL} & \textsc{W20}
    & 25.6
    & 25.3
    & 24.9
    & 24.5
    & \textbf{26.0}
    & 25.2
    & \textbf{26.0}
    & 25.9
    & 25.5
    & \textbf{26.0}
    & 25.9
    & 25.0
    & 25.0
    
    \\
    & \textsc{Flo.}
    & \textbf{22.7}
    & 22.1
    & 21.8
    & 21.6
    & 22.5
    & 22.0
    & 21.8
    & 22.4
    & 22.2
    & 22.0
    & 22.3
    & 21.7
    & 21.8
    \\
    
    \midrule

    EN-PT & \textsc{Flo.}
    & 51.1
    & 50.2
    & 49.8
    & 50.1
    & 50.4
    & 50.1
    & \textbf{51.4}
    & 51.2
    & 49.9
    & 50.5
    & 50.7
    & 49.4
    & 50.4 \\

    \bottomrule

    \end{tabular}
    \caption{BLEU scores on evaluation sets. The top row indicates the training data and its filtering level. \sentences is all of our data filtered through a bitext-filtering pipeline where as \contexts, \loose, \medium, and \strict only include the top 100\%, 75\%, 50\%, and 25\% of documents scored under a SLIDE–CometKiwi filtering metric (Section~\ref{sec:document-filtering}). We indicate whether sentences were concatenated (contextual) or isolated (sentences) during training. We similarly indicate inference input.
    }
    \label{table:bleu-results}
    \vspace{-2mm}
\end{table*}

\subsection{Baselines}
\label{sec:baselines}
We consider two types of sentence-level models for baselines.
For each data filtering level, we additionally train sentence-level models without concatenation.
This allows for comparison on models trained on the same quantity and distribution of data.
Additionally, we produce a \textit{new} dataset that is filtered at the \textit{sentence level} instead of the document level.
Using the entirety of the original data, we filter using traditional bitext filtering pipeline. After deduplication, we remove sentences with: empty lines, more than fifty-percent punctuation, irregular frequencies of characters based on language histograms \cite{DBLP:journals/corr/abs-2010-11125},
uneven length ratios (greater than $1.5$) of source-target, less than 0.5 of target language according to NLLB's fasttext and langid.py, and those below a 0.85 LASER score \cite{schwenk-douze-2017-learning}.
The resulting dataset is described as \textsc{Sentences} in Table~\ref{tab:data-stats}.

\subsection{Training}
\label{sec:training}

All models trained in this paper are trained with the Marian NMT \cite{mariannmt} toolkit.
We train context-aware models with a simple concatenation strategy \cite{tiedemann-scherrer-2017-neural}.
We also train on both contexts and sentences during training via \textsc{sotastream} which mixes two data streams during training \cite{post2023sotastream}.
The first stream which samples from the \paradocs data, concatenates documents up to ten sentences or $256$ tokens (whichever is lesser) and inserts an \texttt{<eos>} token to separate segments.
The second stream pulls from a supplementary dataset composed of preprocessed sentence-level bitext.
These streams are mixed with a 1:1 ratio.
These models are trained with typical next-token prediction and no further alterations.

The models are Transformers \cite{NIPS2017_3f5ee243} with $12$ encoder layers and $6$ decoder layers.
We use a feed-forward dimension of $16,384$ and an embedding size of $1024$.
We train a single \texttt{sentencepiece} vocabulary \cite{kudo-richardson-2018-sentencepiece} for each language pair trained on the supplementary data with a shared vocabulary size of 64k.
All model iterations uses the corresponding vocabulary.
The effective batch size is 500k tokens and one logical epoch is 1B tokens.
We train for 10 logical epochs for a total of 10B tokens/20k updates.
For evaluation, we use the model with the lowest cross-entropy loss per token on the \textsc{Flores200} dev set \cite{nllb2022,flores101,flores}.

\paragraph{Supplementary Data}\label{sec:supplementary-data} is used in addition to the our contextual data.
This ensures that the model can translate both sentences and documents, and is not burdened by a lack of language coverage from high quality sentence pairs mined via other techniques (i.e., global mining that produces high quality datasets such as CCMatrix).
This supplementary data is controlled across experiments.
We filter this bitext in the same way we filter the \sentences baseline described in Section~\ref{sec:baselines}.

\begin{table*}[ht]
        \small
        \centering
\setlength{\tabcolsep}{6pt}
    \begin{tabular}{ll rrrrr rrrr rrrr}
    \toprule

    & & \multicolumn{1}{c}{\textsc{Sent.}}
    & \multicolumn{3}{c}{\contexts}
    & \multicolumn{3}{c}{\loose}
    & \multicolumn{3}{c}{\medium}
    & \multicolumn{3}{c}{\strict} \\

    \cmidrule(lr){3-3} 
    \cmidrule(lr){4-6} \cmidrule(lr){7-9} \cmidrule(lr){10-12} \cmidrule(lr){13-15}

    \multicolumn{2}{c}{Training Type:}
    & \multicolumn{1}{c}{snt.} 
    & \multicolumn{1}{c}{snt.}  & \multicolumn{2}{c}{context} 
    & \multicolumn{1}{c}{snt.}  & \multicolumn{2}{c}{context}
    & \multicolumn{1}{c}{snt.}  & \multicolumn{2}{c}{context}
    & \multicolumn{1}{c}{snt.}  & \multicolumn{2}{c}{context} \\
    \cmidrule(lr){3-3}
    \cmidrule(lr){4-4}
    \cmidrule(lr){5-6}
    \cmidrule(lr){7-7}
    \cmidrule(lr){8-9}
    \cmidrule(lr){10-10}
    \cmidrule(lr){11-12}
    \cmidrule(lr){13-13}
    \cmidrule(lr){14-15}

    \multicolumn{2}{c}{Inference Type:}
    & \multicolumn{1}{c}{snt.} 
    & \multicolumn{1}{c}{snt.} & \multicolumn{1}{c}{snt.}  & \multicolumn{1}{c}{ctx.} 
    & \multicolumn{1}{c}{snt.} & \multicolumn{1}{c}{snt.}  & \multicolumn{1}{c}{ctx.}
    & \multicolumn{1}{c}{snt.} & \multicolumn{1}{c}{snt.}  & \multicolumn{1}{c}{ctx.}
    & \multicolumn{1}{c}{snt.} & \multicolumn{1}{c}{snt.}  & \multicolumn{1}{c}{ctx.} \\

    \midrule
    \multirow{2}{*}{EN-DE}
    
    & Gen.
    & 45.5
    & 44.6
    & 43.3
    & 57.7
    & 44.8
    & 45.4
    & 58.1
    & 45.8
    & 44.0
    & 58.5
    & 45.5
    & 40.8
    & \textbf{60.4}
    \\
    
    & For. 
    & 40.4
    & 42.5
    & 41.5
    & 42.2
    & 43.0
    & 41.4
    & 43.6
    & 41.8
    & 41.5
    & 43.3
    & 42.1
    & 41.5
    & \textbf{44.0}
    
    \\
    
    & Aux.
    & 4.4
    & 4.7
    & 3.6
    & 7.5
    & 4.9
    & 4.8
    & 8.3
    & 3.9
    & 3.6
    & 7.2
    & 4.6
    & 4.4
    & \textbf{9.6}
    \\
    
    \midrule

    \multirow{2}{*}{EN-FR}
    & Gen.

    & 39.5
    & 40.4
    & 40.1
    & 48.4
    & 40.1
    & 39.7
    & 48.7
    & 39.7
    & 39.7
    & \textbf{49.7}
    & 39.6
    & 39.6
    & \textbf{49.7}
    
    \\
    
    & For. 

    & 39.7
    & 38.6
    & 37.9
    & 38.9
    & 38.9
    & 39.8
    & \textbf{42.7}
    & 39.9
    & 39.5
    & 42.0
    & 38.8
    & 39.0
    & 39.8

    \\
    
    & Aux.

    & 0.9
    & 0.9
    & 0.9
    & 7.7
    & 0.8
    & 0.9
    & 10.6
    & 0.9
    & 0.8
    & 10.3
    & 1.2
    & 0.9
    & \textbf{11.6}
    \\
    \midrule

    \multirow{2}{*}{EN-ES} & Gen.
    & 39.3
    & 37.8
    & 37.5
    & 39.3
    & 36.9
    & 38.2
    & 37.3
    & 38.4
    & 37.8
    & 35.7
    & 37.9
    & 38.2
    & \textbf{42.9}
    \\

    & For.
    & \textbf{35.2}
    & 34.5
    & 34.2
    & 32.6
    & 34.3
    & 34.5
    & 31.4
    & 34.9
    & 34.5
    & 30.2
    & 34.7
    & 33.8
    & 16.3
    \\

    & Aux.
    & 0.9
    & 1.0
    & 0.8
    & 9.6
    & 1.0
    & 1.0
    & 10.0
    & 1.0
    & 0.9
    & 10.9
    & 1.6
    & 0.9
    & \textbf{12.5}

    \\
    \midrule

    \multirow{2}{*}{EN-IT} & Gen.
    & 53.0
    & 52.6
    & 51.5
    & 58.0
    & 53.2
    & 53.2
    & 59.2
    & 54.7
    & 53.4
    & \textbf{61.2}
    & 53.7
    & 52.7
    & 60.2
    \\

    & For.
    & 35.2
    & 34.8
    & 34.8
    & 33.6
    & \textbf{36.0}
    & 35.9
    & 34.3
    & 35.9
    & 35.8
    & 35.7
    & 35.7
    & 35.1
    & 34.6
    \\

    & Aux.
    & 1.5
    & 1.7
    & 1.4
    & 5.7
    & 1.8
    & 1.7
    & 5.8
    & 1.8
    & 1.8
    & 7.0
    & 1.8
    & 1.7
    & \textbf{9.9}
    \\
    
    \midrule

    \multirow{2}{*}{EN-PL} & Gen.
    & 30.9
    & 31.3
    & 31.2
    & 36.8
    & 32.0
    & 31.3
    & 37.1
    & 32.6
    & 31.4
    & 37.6
    & 31.8
    & 31.0
    & \textbf{38.5}
    \\

    & For.
    & 30.3
    & 30.6
    & 30.9
    & 31.6
    & 31.4
    & 30.7
    & 30.5
    & 31.4
    & 31.4
    & \textbf{32.8}
    & 30.8
    & 29.9
    & 31.9
    \\

    & Aux.
    & 4.2
    & 4.7
    & 4.7
    & 15.5
    & 6.4
    & 5.2
    & \textbf{18.3}
    & 5.3
    & 6.6
    & 17.0
    & 6.2
    & 4.4
    & 12.3
    \\
    
    & Inf.
    & 33.3
    & 37.6
    & 37.0
    & 40.5
    & 37.0
    & 37.5
    & 40.9
    & 39.2
    & 37.1
    & \textbf{41.3}
    & 38.3
    & 36.6
    & 39.9

    \\
    
    \midrule

    \multirow{2}{*}{EN-PT} & Gen.
    & 36.3
    & 37.3
    & 37.5
    & 42.4
    & 37.1
    & 37.8
    & 43.8
    & 38.2
    & 39.2
    & 46.3
    & 39.5
    & 37.7
    & \textbf{47.0}
    \\

    & For.
    & 21.1
    & 22.7
    & \textbf{22.5}
    & 20.8
    & 22.2
    & 22.3
    & 21.1
    & 21.7
    & 22.2
    & 20.7
    & 21.9
    & 22.1
    & 20.5
    \\

    & Aux.
    & 1.4
    & 2.3
    & 1.5
    & 18.5
    & 3.8
    & 1.6
    & 18.9
    & 3.7
    & 1.6
    & \textbf{24.7}
    & 4.6
    & 1.4
    & 23.3
    \\

    \bottomrule

    \end{tabular}
    \caption{\textsc{ctxpro} scores on evaluation sets. The top row indicates the training data and its filtering level. \sentences is all of our data filtered through a bitext-filtering pipeline where as \contexts, \loose, \medium, and \strict only include the top 100\%, 75\%, 50\%, and 25\% of documents scored under a SLIDE–CometKiwi filtering metric (Section~\ref{sec:document-filtering}). We indicate whether sentences were concatenated (contextual) or isolated (sentences) during training. We similarly indicate inference input.
    }
    \label{table:ctxpro-results}
    \vspace{-2mm}
\end{table*}

\subsection{Evaluation}
\label{sec:evaluation}

We evaluate two-fold: sentence-level translation quality and the ability to address context-dependent phenomena.
The former can be addressed by regular machine translation test sets.
We choose to use \textsc{Flores200} test sets \cite{nllb2022,flores101,flores} as they are available in all of the languages in consideration.
We additionally evaluate on the most recent WMT test set available for that language pair.
For these evaluation sets, we report BLEU \cite{papineni-etal-2002-bleu} scored using \textsc{sacreBLEU} \cite{post-2018-call}.
We additionally report COMET \cite{rei-etal-2020-comet} in Appendix~\ref{sec:appendix} but acknowledge that those scores may be inflated from the COMET-based filtering method.

In order to evaluate the translation of context-dependent phenomena, we turn to \textsc{ctxpro} \cite{wicks-post-2023-identifying} which releases evaluation sets in these languages to evaluate three phenomena: gender, formality, and auxiliary verbs.
Briefly, these handle the translations of English `it' to gendered languages; the translations of English `you' to languages with a T-V distinction; and the translation of English auxiliary verbs in elided sentences such as `I do.'
These are impossible to consistently translate correctly without context.
We report accuracy scores as calculated by the \textsc{ctxpro} PyPI package. 

To translate, we employ two inference designs.
First, for all models we translate using single sentences as input.
Second, for the contextual models, we additionally evaluate them under contextual inference.
For each input, we concatenate up to ten sentences or $256$ tokens (whichever is lesser) of preceding context.
This parallels the contextual data during training.
The last segment is then split (determined by \texttt{<eos>}) and used as the predicted translation.

\section{Results}
\label{sec:results}

In general, we find that context-aware models trained with the \paradocs data outperform
any model trained without contextual inputs across evaluation metrics.
Further, we find that the document-level filtering method is able to improve performance even as data quantity diminishes---indicating a higher quality of data selection.
In Table~\ref{table:bleu-results}, we display BLEU scores as an evaluation of general translation ability and subsequently in Table~\ref{table:ctxpro-results}, we display \textsc{ctxpro} accuracy scores as an evaluation of the ability to translate context-dependent phenomena.
We additionally perform paired bootstrap resampling statistical tests on the evaluation sets to understand whether these differences are meaningful ($p < 0.05$) and comment where appropriate \cite{koehn-2004-statistical}.

\subsection{Effects of Document-Level Filtering}

In Table~\ref{table:bleu-results},
we show BLEU performance of all models.
We compare the contextual models with the sentential baselines.
Within each inference setting, performance steadily, and consistently improves between the models trained on the noisiest data (\contexts) and the cleanest (\strict).
This is true even when \textit{not} using the contextual information available as shown in the sentence-level inference setting of the contextual models.
This is reinforced by the same trend in \textsc{ctxpro} scores displayed in Table~\ref{table:ctxpro-results}.
We note for BLEU scores, there is a small plateau or deterioration between the \medium and \strict data filtering stages and suspect this is a critical trade off point between having high-quantity, high-quality sentence translations (that are \textit{poorer}-quality contextual translations) and having low-quantity, high-quality contextual translations.
This is supported by the fact that we do not see this trend across the \textsc{ctxpro} scores.

\subsection{Sentence-Level Translation Performance}

We also find that by training a model with context-awareness, the model does not lose the ability to translate stand-alone sentences and in many cases, marginally improves.
This is evident by comparing the the models trained under each paradigm (Sentence, and Context Training), when only given sentences during \textit{inference}.
In Table~\ref{table:bleu-results}, this is evident as we see little-to-no difference between the models trained with and without context when they translate individual sentences.
In some cases, we see marginal improvement from the model trained with context.
This indicates that training a machine translation within a context-aware paradigm is \textit{no worse than} training one without.

\subsection{Context-Awareness Boosts BLEU Performance}

Further, we identify benefit from leveraging additional context to translate, even when translating datasets which are not dense with context-dependent phenomena, represented by WMT test sets and Flores200.
Contextual models that are given preceding context during inference get consistent small gains in BLEU compared to their analogous sentence-level models in Table~\ref{table:bleu-results}.
In the larger language pairs (e.g., German and French), we additionally found that the top performing models were statistically outperforming \textit{all} of the sentence-level models however the statistical differences degraded with the smaller language pairs.

\subsection{Translation of Context-Based Phenomena}

Finally, we show that these context-trained models are \textit{more effective} at translating context-dependent phenomena than their sentence-level counterparts.
When considering both Gender and Auxiliaries, the improvement is evident.
We do find the translation of Auxiliaries is still unfortunately low, but may be due to a relatively \textit{low} rare number of occurrences in the dataset.
We also see that in the case of formality, performance is marginally better, though relatively unaffected.
We hypothesize this is due to the particularities of formality translation, rather than exemplification of the dataset, though acknowledge that alternatives may be necessary to target this particular phenomena.
When investigating the statistical differences, we note specifically that in all cases, the context-aware models are statistically superior than their sentence-level counterparts specifically for the auxiliary class.
This is particularly noteworthy as the auxiliary task is the hardest task: the potential correct answer is open to the set of all verbs rather than a simple one-of-three-genders problem as we see with formality.
We also note there is less statistical support for the formality class.
This speaks to the difficulty of this problem and may instead indicate that data is not the simplest solution to approach formality translation.

\section{Analysis of Document Filtering}

Document-level filtering is rare, and to our knowledge, has not been studied.
We demonstrate that our SLIDE–CometKiwi filtering improves contextual performance, but this may be an indication of better sentence-level translations and not better document-level translations.
For a document-level filtering method to hold up against web-crawled data, it should not only filter out poor quality translations, but should also filter out context-inconsistent translations.
By context-inconsistent, we refer to translations that were produced at the sentence-level.
This is roughly synonymous with machine translated text in a web-crawl setting; however, there is no robust machine translation detection methodology to the best of our knowledge.

We instead turn towards a proxy measure to determine if context-inconsistent translations are filtered out.
After scoring each document as described in Section~\ref{sec:document-filtering}, we can rank documents sequentially based off these SLIDE-CometKiwi scores.
We should find more context-consistent translations in the upper percentiles of this distribution and fewer at the lower-end.

\begin{table}[h]
        \centering
    \begin{tabular}{l rrrr}
    \toprule
    & \multicolumn{4}{c}{\textsc{Quartiles}} \\
    & \nth{1} & \nth{2} & \nth{3} & \nth{4}\\
    \midrule
    Inter. Fem. & 46\% & 30\% & 17\% & 6.4\% \\
    Inter. Masc. & 33\% & 41\% & 20\% & 5.8\% \\
    Inter. Neut. & 30\% & 36\% & 24\% & 10\% \\
    \midrule
    Intra. Fem. & 54\% & 29\% & 13\% & 3.5\% \\
    Intra. Masc. & 39\% & 39\% & 18\% & 4.5\% \\
    Intra. Neut & 35\% & 35\% & 21\% & 9.4\% \\
\bottomrule
\end{tabular}
    \caption{For each category of gendered pronoun, what percent of all examples occur in each quartile defined by SLIDE–CometKiwi scores. Intersential indicates the antecedent occured in the same sentence while intrasential indicates it occured in a different sentence.}
    \label{tab:ctxpro-annotations}
    \vspace{-2mm}
\end{table}

The \texttt{ctxpro} toolkit, which generated these evaluation sets, also identifies when there are ambiguous uses of gendered pronouns.
As the toolkit is intended to be high precision, low recall, it is extremely likely to not identify ambiguity in pronouns when the translations are incorrect or inconsistent.
Thus, we can use the \texttt{ctxpro} toolkit as a proxy measure for context-consistent translations.

We identify all examples of gendered-pronouns in our data according to the \texttt{ctxpro} toolkit, and investigate where SLIDE-CometKiwi ranks them in relation to the remaining documents.
We show percentages in Table~\ref{tab:ctxpro-annotations}.
We discriminate between intersentential---where the antecedent is in the same sentence---and intrasentential---where the antecedent is in a previous sentence.
The former is quite easy to correctly translate for a sentence-level model while the latter is impossible.
We further distinguish by the gender of the pronoun---where neuter is the majority class.

As shown in Table~\ref{tab:ctxpro-annotations}, most of the identified examples occur in the \nth{1} or \nth{2} quartile.
Further, we see that this is \textit{especially} true for the feminine pronouns (a quintessential minority class for pronouns).
Conversely, neuter pronouns have a more uniform distribution across quartiles.
We hypothesize this indicates there is more machine translated texts in the lower quartiles---as they would have incidentally correctly translated neuter pronouns---and less in the top.
More specifically, the \textit{most challenging} translation for machines, ambiguous pronouns with an intrasentential feminine antecedent, are mostly ranked in the top quartile.
We assume this also means the top quartile was translated by humans.

\section{Conclusions}

Research in contextual machine translation is hampered by the lack of document annotations on parallel data.
We augment three large popular MT datasets (ParaCrawl, News Commentary, and Europarl) with this information, creating a document-level dataset, \paradocs.
We introduce a document-level filtering method to apply to this data in lieu of traditional context-destroying sentence filtering methods.
Simple, context-aware machine translation models trained on this data have shown to be better at machine translation in both general performance as measured by WMT test sets as well as targeted performance---measured by the ability to correctly translate discourse phenomena.
We release the data as a resource to the community.\footnote{https://huggingface.co/datasets/jhu-clsp/paradocs}

\section{Limitations}
\label{sec:limitations}
As mentioned in this work, document reconstruction is particularly constrained by the original dataset processing.
Many languages have only been processed with the global mining technique (i.e., CCMatrix) and ParaCrawl notably only supports European languages.
This works also assumes there exists many well-translated parallel documents in web-crawled corpora.
Not only is this not true for many language pairs, but there is recent evidence to suggest that the more multi-way parallel data is, the more likely it was machine translated \cite{thompson2024shocking}.

This also significantly constrains the amount of data.
In our smallest setting, English–German (an extraordinarily high-resource language pair) is still limited to 42M lines.
If we were to extend this to lower-resource languages, we would be limited to perhaps a few thousand lines which are unlikely to make any meaningful difference in performance.

\section{Acknowledgements}
We would like to thank the reviewers for their discussion and feedback. We additionally would like to thank Barry Haddow, Marc Marone, Neha Verma, and Joe McKnight for their resources, technical discussions, and support.

\bibliography{anthology, custom}
\bibliographystyle{acl_natbib}
\newpage
\appendix
\section{Appendix}
\label{sec:appendix}

\begin{table}[ht]
        \centering
\setlength{\tabcolsep}{6pt}
    \begin{tabular}{l rr r}
    \toprule
    Dataset & \# Lines \\
    \midrule
    OPUS-ccaligned-v1-deu-eng & 9.2M \\ %
    OPUS-ccmatrix-v1-deu-eng & 167M \\ %
    OPUS-news\_commentary-v16-deu-eng & 222k \\%222 215
    OPUS-paracrawl-v9-deu-eng & 140M \\%139 679 498
    OPUS-wikimatrix-v1-deu-eng & 1.1M \\ %
    OPUS-wmt\_news-v2019-deu-eng & 35k \\ %
    \midrule
    total & 317M \\ %
    \bottomrule

    \end{tabular}
    \caption{English--German supplementary sentence-level bitext. Names based on the published \texttt{mtdata} name. \url{https://github.com/thammegowda/mtdata}. Flores200 dev and devtest specifically removed from this data before training.}
    \label{table:data}
\end{table}

\begin{table}[ht]
        \centering
\setlength{\tabcolsep}{6pt}
    \begin{tabular}{l rr r}
    \toprule
    Dataset & \# Lines \\
    \midrule

    OPUS-ccaligned-v1-eng-spa & 8.6M \\ %
    OPUS-ccmatrix-v1-eng-spa & 345M\\ %
    OPUS-elrc$^{*}$ & 1.5M \\ %
    OPUS-europarl-v8-eng-spa & 1.7M\\ %
    OPUS-multiccaligned-v1-eng-spa & 30M \\ %
    OPUS-multiparacrawl-v7.1-eng-spa & 54M\\ %
    OPUS-news\_commentary-v16-eng-spa & 38k \\ %
    OPUS-paracrawl-v9-eng-spa & 154M \\ %
    OPUS-ted2020-v1-eng-spa & 299k \\ %
    OPUS-wmt\_news-v2019-eng-spa & 11k\\ %
    OPUS-wikimatrix-v1-eng-spa & 2.6M \\ %
    OPUS-wikipedia-v1.0-eng-spa & 1.3M \\ %
    OPUS-wikimedia-v20210402-eng-spa & 871k \\ %
    \midrule
    total & 600M \\ %
    \bottomrule

    \end{tabular}
    \caption{English--Spanish supplementary sentence-level bitext. Names based on the published \texttt{mtdata} name. \url{https://github.com/thammegowda/mtdata}. Flores200 dev and devtest specifically removed from this data before training.}
    \label{table:data}

\end{table}

\begin{table}[ht]
        \centering
\setlength{\tabcolsep}{6pt}
    \begin{tabular}{l rr r}
    \toprule
    Dataset & \# Lines \\
    \midrule
    OPUS-ccaligned-v1-eng-fra & 9.4M \\%9 407 935 
    OPUS-ccmatrix-v1-eng-fra & 263M \\ %
    OPUS-elrc$^{*}$ & 2.9M \\ %
    OPUS-europarl-v8-eng-fra & 1.8M \\ %
    OPUS-news\_commentary-v14-eng-fra & 118k \\%118 656 \\
    OPUS-paracrawl-v9-eng-fra & 137M \\%137 222 460 \\
    OPUS-ted2020-v1-eng-fra & 276k \\%275 369 \\
    OPUS-wikimatrix-v1-eng-fra & 2.2M \\%2 178 543 \\
    OPUS-wikimedia-v20210402-eng-fra & 657k \\ %
    OPUS-wikipedia-v1 & 417k \\ %
    OPUS-wmt\_news-v2019-eng-fra & 19k \\%19 017 \\
    \midrule
    total & 418M \\ %
    \bottomrule

    \end{tabular}
    \caption{English--French supplementary sentence-level bitext. Names based on the published \texttt{mtdata} name. \url{https://github.com/thammegowda/mtdata}. Flores200 dev and devtest specifically removed from this data before training.}
    \label{table:data}
\end{table}

\begin{table}[ht]
        \centering
\setlength{\tabcolsep}{6pt}
    \begin{tabular}{l rr r}
    \toprule
    Dataset & \# Lines \\
    \midrule
OPUS-ccaligned-v1-eng-ita & 8.9M \\ %
OPUS-ccmatrix-v1-eng-ita & 121k\\ %
OPUS-elrc$^{*}$ & 1.0M \\ %
OPUS-news\_commentary-v16-eng-ita & 67k \\ %
OPUS-multiccaligned-v1-eng-ita & 20.4M \\ %
OPUS-paracrawl-v9-eng-ita & 6.1M \\ %
OPUS-wmt\_news-v2019-eng-ita & 2.4k \\ %
OPUS-wikimatrix-v1-eng-ita & 1.7M \\ %
OPUS-wikipedia-v1.0-eng-ita & 307k \\ %
OPUS-wikimedia-v20210402-eng-ita & 232k \\ %
\midrule
total & 215M \\ %
    \bottomrule

    \end{tabular}
    \caption{English--Italian supplementary sentence-level bitext. Names based on the published \texttt{mtdata} name. \url{https://github.com/thammegowda/mtdata}. Flores200 dev and devtest specifically removed from this data before training.}
    \label{table:data}

\end{table}

\begin{table}[ht]
        \centering
\setlength{\tabcolsep}{6pt}
    \begin{tabular}{l rr r}
    \toprule
    Dataset & \# Lines \\
    \midrule
OPUS-ccaligned-v1-eng-pol & 6.1M \\ %
OPUS-ccmatrix-v1-eng-pol & 50M \\ %
OPUS-elrc$^{*}$ & 1.3M \\ %
OPUS-europarl-v8-eng-pol & 520k \\ %
OPUS-multiccaligned-v1-eng-pol & 8.1M \\ %
OPUS-multiparacrawl-v7.1-eng-pol & 8.7M \\ %
OPUS-paracrawl-v9-eng-pol & 21.9M \\ %
OPUS-ted2020-v1-eng-pol & 107k \\ %
OPUS-wikimatrix-v1-eng-pol & 370k \\ %
OPUS-wikipedia-v1.0-eng-pol & 100k \\ %
OPUS-wikimedia-v20210402-eng-pol & 31k \\ %
OPUS-elra$^{*}$ & 122k\\ %
OPUS-kde4-v2-eng-pol & 37k \\ %
OPUS-dgt-v2019-eng-pol & 2.1M \\ %
\midrule
total & 102M \\ %
    \bottomrule

    \end{tabular}
    \caption{English--Polish supplementary sentence-level bitext. Names based on the published \texttt{mtdata} name. \url{https://github.com/thammegowda/mtdata}. Flores200 dev and devtest specifically removed from this data before training.}
    \label{table:data}

\end{table}

\begin{table}[ht]
        \centering
\setlength{\tabcolsep}{6pt}
    \begin{tabular}{l rr r}
    \toprule
    Dataset & \# Lines \\
    \midrule
OPUS-ccaligned-v1-eng-por & 7.3M \\ %
OPUS-ccmatrix-v1-eng-por & 147M \\ %
OPUS-elrc$^{*}$ & 1.4M \\ %
OPUS-europarl-v8-eng-por & 1.7M \\ %
OPUS-multiccaligned-v1-eng-por & 13M \\ %
OPUS-multiparacrawl-v7.1-eng-por & 22M\\ %
OPUS-news\_commentary-v16-eng-por & 48k\\ %
OPUS-paracrawl-v9-eng-por & 54M \\ %
OPUS-ted2020-v1-eng-por & 227k \\ %
OPUS-wikimatrix-v1-eng-por & 2.0M \\ %
OPUS-wikipedia-v1.0-eng-por & 1.0M \\ %
OPUS-wikimedia-v20210402-eng-por & 363k \\ %
    \midrule
    total & 253M \\ %
    \bottomrule

    \end{tabular}
    \caption{English--Portuguese supplementary sentence-level bitext. Names based on the published \texttt{mtdata} name. \url{https://github.com/thammegowda/mtdata}. Flores200 dev and devtest specifically removed from this data before training.}
    \label{table:data}

\end{table}

\begin{table*}[ht]
        \small
        \centering
\setlength{\tabcolsep}{6pt}
    \begin{tabular}{ll rrrrr rrrr rrrr}
    \toprule

    &   
    & \multicolumn{1}{c}{\textsc{Sent.}}
    & \multicolumn{3}{c}{\contexts}
    & \multicolumn{3}{c}{\loose}
    & \multicolumn{3}{c}{\medium}
    & \multicolumn{3}{c}{\strict} \\

    \cmidrule(lr){3-3} 
    \cmidrule(lr){4-6} \cmidrule(lr){7-9} \cmidrule(lr){10-12} \cmidrule(lr){13-15} 

    \multicolumn{2}{c}{Training Type:}
    & \multicolumn{1}{c}{snt.} 
    & \multicolumn{1}{c}{snt.}  & \multicolumn{2}{c}{context} 
    & \multicolumn{1}{c}{snt.}  & \multicolumn{2}{c}{context}
    & \multicolumn{1}{c}{snt.}  & \multicolumn{2}{c}{context}
    & \multicolumn{1}{c}{snt.}  & \multicolumn{2}{c}{context} \\
    \cmidrule(lr){3-3}
    \cmidrule(lr){4-4}
    \cmidrule(lr){5-6}
    \cmidrule(lr){7-7}
    \cmidrule(lr){8-9}
    \cmidrule(lr){10-10}
    \cmidrule(lr){11-12}
    \cmidrule(lr){13-13}
    \cmidrule(lr){14-15}

    \multicolumn{2}{c}{Inference Type:}
    & \multicolumn{1}{c}{snt.} 
    & \multicolumn{1}{c}{snt.} & \multicolumn{1}{c}{snt.}  & \multicolumn{1}{c}{ctx.} 
    & \multicolumn{1}{c}{snt.} & \multicolumn{1}{c}{snt.}  & \multicolumn{1}{c}{ctx.}
    & \multicolumn{1}{c}{snt.} & \multicolumn{1}{c}{snt.}  & \multicolumn{1}{c}{ctx.}
    & \multicolumn{1}{c}{snt.} & \multicolumn{1}{c}{snt.}  & \multicolumn{1}{c}{ctx.} \\

    \midrule
    \multirow{2}{*}{EN-DE} & \textsc{W23}
    & 79.9
    & 80.4
    & 81.4
    & 81.5
    & 81.7
    & 80.1
    & 81.0
    & 81.5
    & 81.1
    & 80.5
    & 81.7
    & \textbf{82.2}
    & 81.8
    \\
    & \textsc{Flo.}
    & 87.2
    & 87.2
    & 87.8
    & 87.7
    & \textbf{88.1}
    & 86.9
    & 87.5
    & 88.0
    & 87.7
    & 87.7
    & 87.9
    & 87.9
    & 87.8
    \\
    \midrule

    \multirow{2}{*}{EN-ES} & \textsc{W13}
    & 86.1
    & 86.2
    & 86.0
    & 86.1
    & 86.1
    & 86.1
    & 86.1
    & \textbf{86.3}
    & \textbf{86.3}
    & 86.1
    & \textbf{86.3}
    & 86.2
    & 86.1
    \\
    & \textsc{Flo.}
    & 86.0
    & 86.0
    & 86.0
    & 86.0
    & 86.2
    & 86.1
    & 86.2
    & 86.3
    & 86.3
    & 86.1
    & \textbf{86.5}
    & 86.3
    & 86.2
    \\
    \midrule

    \multirow{2}{*}{EN-FR} & \textsc{W15}
    
    & 83.5
    & 83.8
    & 84.0
    & 84.2
    & 84.1
    & 83.6
    & 84.1
    & \textbf{84.3}
    & 83.9
    & 83.1
    & 83.4
    & 84.1
    & 83.5
    
    \\
    & \textsc{Flo.}
    
    & 88.0
    & 87.7
    & 88.0
    & 88.2
    & 88.4
    & 87.9
    & 88.2
    & \textbf{88.5}
    & 88.1
    & 88.0
    & 88.0
    & 88.1
    & 88.0
    
    \\
    \midrule
    \multirow{2}{*}{EN-IT} & \textsc{W09}
    & 86.6
    & 86.5
    & 86.5
    & 86.5
    & 86.9
    & 86.7
    & 86.9
    & \textbf{87.1}
    & 86.9
    & 86.9
    & 86.9
    & 86.6
    & 86.6
    \\
    & \textsc{Flo.}
    & 87.7
    & 87.4
    & 87.3
    & 87.0
    & 87.8
    & 87.8
    & 87.6
    & \textbf{88.2}
    & 88.0
    & 87.9
    & 88.0
    & 87.8
    & 87.6

    \\
    \midrule
    
    \multirow{2}{*}{EN-PL} & \textsc{W20}
    & 86.7
    & 86.3
    & 85.5
    & 83.8
    & \textbf{87.3}
    & 86.4
    & 86.5
    & 87.1
    & 86.8
    & 86.9
    & 86.9
    & 86.3
    & 86.3
    \\
    & \textsc{Flo.}
    & 88.0
    & 87.4
    & 87.0
    & 86.9
    & 88.1
    & 87.8
    & 87.8
    & \textbf{88.4}
    & 88.1
    & 88.0
    & 88.2
    & 87.8
    & 87.6
    \\
    \midrule

    \multirow{1}{*}{EN-PT}
    & \textsc{Flo.}
    & 89.6
    & 89.1
    & 89.0
    & 89.1
    & 89.3
    & 89.4
    & 89.7
    & \textbf{89.8}
    & 89.5
    & 89.7
    & \textbf{89.8}
    & 89.4
    & 89.4
    \\
    
    \bottomrule

    \end{tabular}
    \caption{COMET scores (x100) on evaluation sets. The top row indicates the training data and its filtering level. \sentences is all of our data filtered through a bitext-filtering pipeline where as \contexts, \loose, \medium, and \strict only include the top 100\%, 75\%, 50\%, and 25\% of documents scored under a SLIDE–CometKiwi filtering metric (Section~\ref{sec:document-filtering}). We indicate whether sentences where concatenated (contextual) or isolated (sentences) during training. We similarly indicate inference input.
    }
    \label{table:comet-results}
\end{table*}

\end{document}